\documentclass{AAIML2}

\usepackage{amsmath,amssymb,mathrsfs,bm,enumerate,physics}
\usepackage{autobreak,mathtools}
\usepackage{tikz,float,placeins}
\usetikzlibrary{intersections,calc,arrows.meta}
\usepackage{ulem}
\usepackage{cleveref}
\usepackage{graphicx,booktabs,capt-of}
\usepackage{setspace}
\usepackage{enumitem}

\usepackage{spverbatim}
\usepackage{caption}
\usepackage{array}
\usepackage{adjustbox}
\usepackage{comment}

\usepackage{multirow}
\usepackage{diagbox} 

\newcommand\myshade{75}
\colorlet{mylinkcolor}{red}
\colorlet{mycitecolor}{green}
\colorlet{myurlcolor}{blue}
\hypersetup{colorlinks=true, anchorcolor=cyan, 
linkcolor=mylinkcolor!\myshade!black, 
citecolor=mycitecolor!\myshade!black,
urlcolor=myurlcolor!\myshade!black } 
\newcommand{\beq}{\begin{equation}}
\newcommand{\eeq}{\end{equation}}

\makeatletter
\long\def\@makefntext#1{\parindent 1em\noindent
\@hangfrom{\hbox to 1.8em{\hss$^{\@thefnmark}$}}#1}
\makeatother

\def\blue#1{{\color{black}#1}}

\aaimlheading{vol}{issue}{pp-pp}{dd-mm-yy}{dd-mm-yy}{dd-mm-yy}{Haru-Tada Sato and Fuka Matsuzaki}

\ShortHeadings{Exploring the Limits of LLM}{Sato and Matsuzaki}
\firstpageno{00}

\mycitation{Haru-Tada Sato, et al.}{Exploring the Limits of Large Language Models: A Systematic Evaluation of Masked Text Processing Ability through MskQA and MskCal.}{Advances in Artificial Intelligence and Machine Learnin.}{2025;0(0):0.}{}

\begin{document}

\title{Exploring the Limits of Large Language Models: A Systematic Evaluation of Masked Text Processing Ability through MskQA and MskCal}

\author{\name Haru-Tada Sato \email satoh@isfactory.co.jp \\
\name Fuka Matsuzaki \email matsuzaki@isfactory.co.jp\\
       \addr Department of Data Science\\
       i's Factory Corporation, Ltd.\\
       Kanda-nishiki-cho 2-7-6, Tokyo 101-0054, JAPAN
}

\editor{Haru-Tada Sato}

\maketitle

\begin{abstract}
This paper evaluates Large Language Models' (LLMs) limitations through their masked text processing capabilities. We introduce two novel tasks: MskQA, which measures reasoning on masked question-answering datasets, and MskCal, which assesses numerical reasoning on masked arithmetic problems. Testing GPT-4o and 4o-mini reveals that LLM performance depends significantly on masking rates and semantic information availability. Our experiments demonstrate that performance decreases as semantic information is reduced, with GPT-4o consistently outperforming 4o-mini, particularly in numerical reasoning tasks. The study shows that LLMs maintain reasonable accuracy with masking rates below 40\%, but struggle with heavily masked computational tasks. Our findings illuminate the interaction between background knowledge and reasoning ability in masked text processing, highlighting the need for more robust methods to assess LLMs' true comprehension capabilities.~\footnote{Codes and data will be available at 
https://github.com/maskcode9004/maskcode}
\end{abstract}


\setlength{\parindent}{1em}
\section{Introduction}

Large-scale language models (LLMs)~\cite{Islam2024}-\cite{Touvron2023} have demonstrated remarkable abilities in a variety of tasks, and as the scale of the model increases, certain abilities emerge (often referred to as emergent abilities)~\cite{Wei2022a, Wei2022b}. 
One notable research direction examining such capabilities analyzes LLM's robustness to word and character rearrangements (input perturbations), providing insights into inference mechanisms. The underlying mechanisms behind these abilities remain largely unexplained, and understanding them is expected to clarify the differences between various LLMs and human cognition. 
\blue{
In particular, it remains an open question whether LLMs primarily rely on syntactic structures, semantic associations, or memorized patterns when making inferences 
-- especially under conditions where surface-level information is obscured.}

GPT-4 shows near-perfect performance in processing text heavily scrambled at the character level \cite{Matsuo2023}, demonstrating an ability to recover original sentences even with randomly scrambled characters, similar to humans' typoglycemia recognition capacity. 
Prior studies have investigated word-level reordering effects as well. For example, \cite{Sinha2021a}-\cite{Zhao2024} discuss LLMs' insensitivity to word order disruptions in downstream tasks like summarization and conversation generation. These findings question a fundamental assumption about LLMs, namely that they understand syntax and use it to comprehend sentences in human-like ways.

While our investigation also explores responses to input variations, we focus on non-perturbative inputs rather than rearrangements. In rearrangement tasks, original information persists through variations from initial states, but extreme rearrangements approach non-perturbative operations where meaning fundamentally changes. Our study examines such operations through strategic masking, offering a distinct perspective on input manipulation.

Our verification approach replaces essential question-answering information with meta-information to examine inference accuracy. While superficially similar to Masked Language Model (MLM) evaluation~\cite{Salazar2020,Song2019}, our task is more demanding: we assess not only masked word prediction but also accurate inference performance while text remains masked, thereby evaluating models' grasp of contextual and logical structures~\cite{Sun2019}. We extend our investigation beyond question-answering to mathematical reasoning, examining both multiple-choice questions and free-form responses with guided prompts.

The cryptic decoding task deeply relates to contextual understanding, which appears connected to large-scale language models' learning mechanisms. However, why GPT --- a unidirectional decoder unlike encoder models such as BERT --- exhibits strong contextual understanding remains unexplained. While designed for autoregressive text generation, GPT's capacity to handle tasks requiring deep contextual comprehension suggests 
that underlying mechanisms supporting this ability are not fully understood \cite{OpenAI2023}-\cite{Brown2020}.

\vspace{-0.7cm}
\hspace{-0.09\linewidth}
\begin{figure}[h]	
	\begin{minipage}{0.9\hsize}
        \includegraphics[scale=0.24]{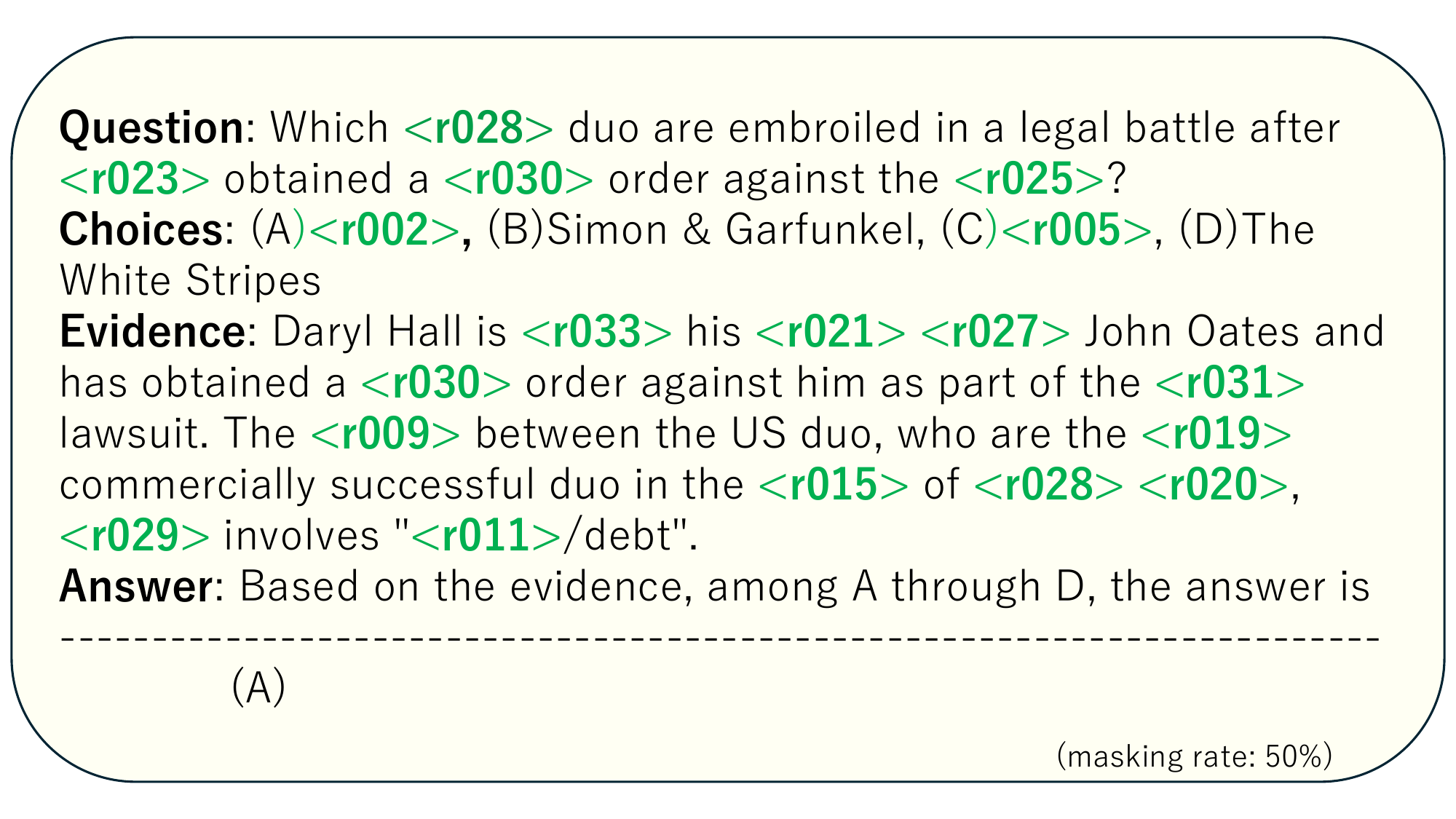}
        \includegraphics[scale=0.24]{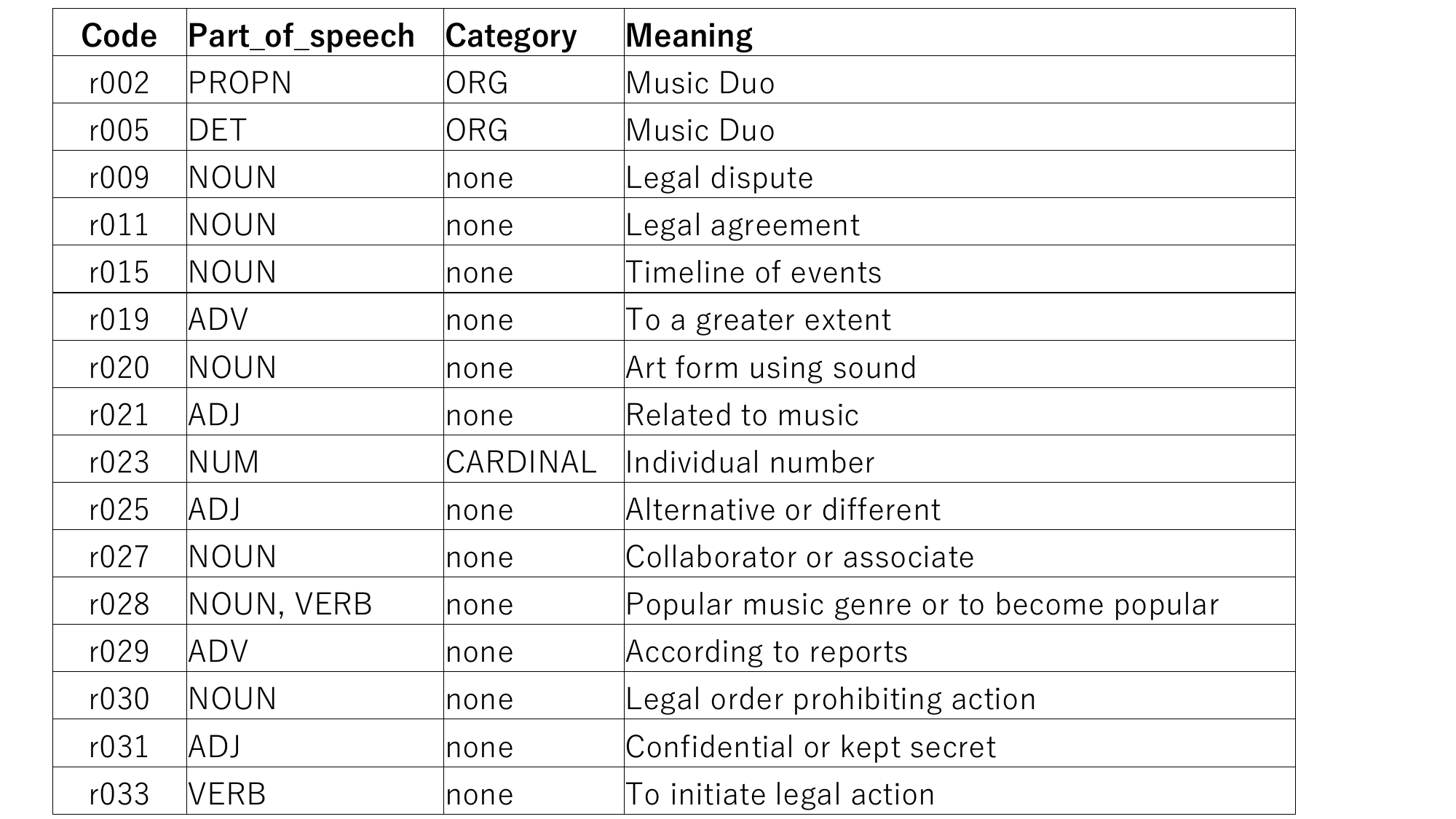}
	\end{minipage}
        \captionsetup{width=0.95\linewidth}
    	\caption{Masked question example. Left: Question with semantic elements replaced by code references. Right: Mask code definitions (part-of-speech, category, meaning).}
     \label{fig:qa_sample}
     \vspace{-0.5cm}
\end{figure}

%
%
\section{Problem formulation}
Figure \ref{fig:qa_sample} provides an example of a masked question, which is masked using a masking code, as illustrated on the right-side. This code is associated with part-of-speech (POS) tags, abstracted category information, and semantic information. The quality and quantity of meta-information significantly impact response accuracy. Qualitative control primarily involves adjusting the semantic information's abstraction level, while quantitative control can employ strict masking --- where no semantic information is provided --- or apply mask codes to specific parts of speech. Using these various masking schemes, we observe how answer accuracy changes as masking rate varies.

While various observation methods are available, evaluation presents challenges since masked text contains abstract content with similarly abstract responses. Therefore, we must establish question formats allowing clear evaluation. To assess LLMs' masked text processing ability, we propose two decoding tasks:

\noindent
(1) {\bf Masked Question Answering (MskQA)}: This task directly measures LLMs' ability to recognize masked concepts and construct meaning by providing questions with evidence and multiple-choice options. Performance is quantitatively evaluated using answer slection accuracy, offering a clear metric for assessing inference ability under masked conditions.

\noindent
(2) {\bf Masked Calculations (MskCal)}: LLMs perform calculations based on guided questions describing calculation procedures. While this task can handle multiple-choice formats like MskQA, we focus on numerical answer generation to evaluate error rates. Since LLMs generate responses based on vocabulary probabilities rather than actual computation, this task specifically assesses their ability to construct computational logic under masked conditions (see MskCal prompt example in Appendix~\ref{sec:C}).

\subsection{Dataset}
\vspace{-1em}
The dataset preparation follows the method outlined by \cite{Matsuo2023}. If test data for downstream tasks is included in the training set, models may simply memorize the original content and provide answers regardless of decoding ability, raising concerns about evaluation fairness. To address this issue, we primarily use a masked version of RealtimeQA \cite{Kasai2022} (RQA) for evaluation. RQA is a dynamic question-answering dataset publishing questions about the latest news weekly, making it unlikely current LLMs have memorized the content.

Specifically, we collect the latest RQA data (November 1, 2023, to January 11, 2024) and process evidence sentences to construct MskQA samples. To prevent learned knowledge use, we focus on data after GPT-4's September 2023 cut-off time. We employed GPT-4o-mini to identify and exclude insufficient evidence and randomly selected 100 samples from the remainder. We removed URL references from evidence, as this information could be directly accessed (comparing accuracy rates before and after URL removal confirmed higher accuracy prior to removal).

For MskCal, we evaluate both multiple-choice and written-answer formats. The multiple-choice format follows MskQA's structure, utilizing a masked version of AQuA-RAT \cite{Ling2017} (AQA)—a dataset of mathematical word problems requiring multi-step reasoning. Instead of evidence sentences, we employ chain-of-thought (CoT) prompts similar to \cite{Wei2022b}. To test whether models can perform appropriate reasoning under masked conditions, we verify three settings: (1) thought process descriptions using identical numerical values as questions but omitting correct answers, (2) thought processes providing full reasoning leading to correct answers while questions use different numerical values, and (3) no thought process provided for comparison. 
To ensure computational reasoning is appropriately evaluated in masked states, neither multiple-choice options nor numerical values affecting calculation results are masked.

The descriptive questions ask simple accounting calculations. The numerical values and prerequisites required for calculation are given as input, and questions are designed to guide the calculation process so inferences can be made by following the procedure. Here too, numerical values, prerequisites, and guidance parts remain unmasked, while remaining parts are masked. For specific prompts, see Appendix \ref{sec:C}.

A higher masking rate is expected to reduce the influence of background knowledge. To quantify this effect, we compare the results of MskQA with data assumed to incorporate background knowledge. Specifically, we use UQA, a set of 100 questions generated by an LLM from universal documents such as the Universal Declaration of Human Rights, as a potential source of background knowledge. Since questions whose answers depend on prior knowledge are unsuitable for this verification, we ensure that all generated questions have answers explicitly present in the evidence text. For details on the generation method, see Appendix~\ref{sec:A}.

\subsection{Masking}
\vspace{-1em}

For each dataset, the masking process generates masked text at multiple masking rates: we randomly select words at a fixed rate and generate masked codes for each selected word. Masking targets content words (nouns, verbs, adjectives, adverbs) while excluding function words (prepositions, articles, conjunctions, auxiliaries, pronouns).

The masking procedure follows these steps: 1) Create a morpheme list (content words), which forms the basis for masking rate calculation; 2) Control masked word quantity through random sampling according to specified masking rate; 3) Given a source sentence and reduced list, ask the language model to create meta-information including mask codes for the masking task.
 
\subsection{Metrics}
\vspace{-1em}

We define accuracy rate $Acc(D)$ as the ratio of correct answers to all questions based on evidence sentences in dataset $D$. Given a dataset $D$ with masking rate $r$, the question set is denoted as $D_r$, where $D_0$ refers to the unmasked dataset. The accuracy for masking rate $r$ is expressed as $Acc(D_r)$. Since $Acc(D_r)$ may depend on question difficulty in $D$, normalization is necessary to remove this dependency. We define normalized accuracy rate $NA(D,r)$ as $\frac{Acc(D_r)}{Acc(D_0)}$. Unless specified otherwise, we use RQA as dataset $D$ and denote question-answering data generated from UQA as $U$.

When masking rate is low, background information can be inferred from textual content. Therefore, we need metrics accounting for this effect. The normalized accuracy $NA(U,r)$, derived from dataset $U$, assumes that the model already possesses relevant background knowledge. Thus, background knowledge dependence can be quantified by comparing results against this baseline. While $Acc(D_0)$ absolute value doesn't directly represent knowledge dependence, its contribution can be assessed by comparing to $NA(U,r)$.

To quantify background knowledge dependence, one approach evaluates the ratio relative to $NA(U,r)$; a smaller ratio indicates lower dependence (greater independence). Alternatively, to equalize background knowledge effect and reevaluate results, one could average $NA(D,r)$ with $NA(U,r)$. For example, the latter approach computes geometric mean 
\[PA(D,r) = \sqrt{NA(D,r) \cdot NA(U,r)}\,, \]
and $Acc(D_0) \cdot PA(D,r)$ yields effective accuracy $EA(D_r)$. When $D=U$, the dataset serves as its own baseline, resulting in $PA(U,r)=NA(U,r)$ and $EA(U,r)=Acc(U_r)$.

The former approach --- designed to increase with decreasing knowledge dependence --- can be interpreted by inverting the sign. Furthermore, if $Acc(U,r) > Acc(D,r)$, the gain effect from acquiring background knowledge can be represented as $Acc(U,r) - Acc(D,r)$. The larger this value, the lower the dependence on background knowledge (higher the independence). To formalize this, we define the normalized gain (termed knowledge independence $KI$) as: 
\beq
 KI(D,r) = 1 - \frac{Acc(D,r)}{Acc(U,r)}\,. \label{KI} 
\eeq
Here, $KI$ indicates that larger values mean greater the knowledge independence (lower external knowledge dependence). However, if the assumption $Acc(U,r) \geq Acc(D,r)$ is violated, $KI(D,r)$ may become negative. When $D = U$, the knowledge independence score becomes zero, i.e., $KI(r) = 0$.

\section{Variation of masking methods}\label{sec:mr}

As described in the problem setting, the objective of this experiment is to examine LM response accuracy in relation to masking rate. The masking task is performed by a different LM than the one used for decoding. This separation primarily eliminates LM-specific idiosyncrasies that could arise if the same LM performed both tasks, including mitigating memorization risks. Since morpheme and vocabulary identification capabilities vary between LMs, maintaining consistent masking coverage may be difficult. To ensure consistent masking control, we supplement LM-based masking with Python-based processing.

Considering cost and budgetary constraints, we employ Gemma2, a lightweight LM offering affordability and fast processing for the masking task. For the decoding task, we compare multiple models to observe performance variations across large-scale and lightweight LMs.
\subsection{Handling missing information}
\vspace{-1em}
We define masking rate (MR) as the ratio of masked words to total maskable words. Here, "number of words" refers to morpheme count, with compound words counted as single words.

Decoding task accuracy is influenced not only by the decoding model's precision but also by masking task quality. Naturally, accuracy decreases in the "{\bf solid masking state}," where masked content lacks any meta-information despite conversion to mask codes. The masking task inevitably generates a certain number of failures in producing meaningful information (e.g., abstracting meaning only to category information or failing to produce output due to system processing issues). Therefore, understanding the impact when significant portions of meaning information are not generated becomes crucial.

To assess this impact, we compare decoding under two conditions: one where some portion of masking codes includes solid masking, and one where all solid masking codes are lifted. Since the masking task inherently contains system-generated missing information, we refer to the former as {\bf regular masking} (where solid masking is retained), and the latter as {\bf partial lifting} (where all solid masking codes are restored to their original state without masking). This comparison allows us to evaluate solid masking's impact on decoding performance.

\subsection{Impact of meta-information}\label{sec:black}
\vspace{-1em}

\blue{
The mask codes are generated through a prompt-based process using a lightweight language model (Gemma2), applied to POS-tagged content words extracted via a morpheme parser. Each masked token is replaced with a structured code that includes part-of-speech, semantic category, and abstract meaning. To ensure consistency and reproducibility, a fixed prompt template was used, and morpheme coverage was verified using deterministic token lists.}

Accuracy may still be affected by meta-information quality provided to mask codes. If meta-information contains sparse semantic information, it approaches blackout status, decreasing accuracy. Conversely, dense information should increase inference accuracy. We investigate how qualitative differences in meta-information reflect in MR-Acc response curves, particularly whether partial lifting curve position appears above the regular position.

We employ two additional masking techniques for comparison beyond regular masking and partial lifting: strict masking, where semantic information is completely blacked out (though morpheme information and categories are provided, regardless of whether the category is none or empty); and lenient masking, which does not mask verbs. If words with the same lemma appear in non-verb POS categories, these words are also excluded from masking. Since our main focus is examining differences in masking task processing impact, we use a simple method.

We employ two additional masking techniques for comparison beyond regular masking and partial lifting: {\bf strict masking}, where semantic information is completely blacked out (though morpheme information and categories are provided, regardless of whether the category is none or empty); and {\bf lenient masking}, which does not mask verbs. If words with the same lemma appear in non-verb POS categories, these words are also excluded from masking. Since our main focus is examining differences in masking task processing impact, we use a simple method.


%
\section{Results}

\blue{
We note that the results involve multiple masking strategies, datasets, and models, making the structure inherently complex. While we summarize task definitions and masking strategy variations in text, future versions of this framework may benefit from including visual flowcharts or comparative tables to further clarify the processing pipeline. 
In the meantime, readers may refer to the publicly available code and sample prompts on our GitHub repository, which illustrate the masking and decoding workflows more concretely.}

\subsection{RQA and UQA}
\vspace{-1em}

\begin{figure}[h]
\hspace{-0.07\linewidth}
\begin{minipage}[t]{0.62\linewidth}
    \includegraphics[width=0.95\linewidth]{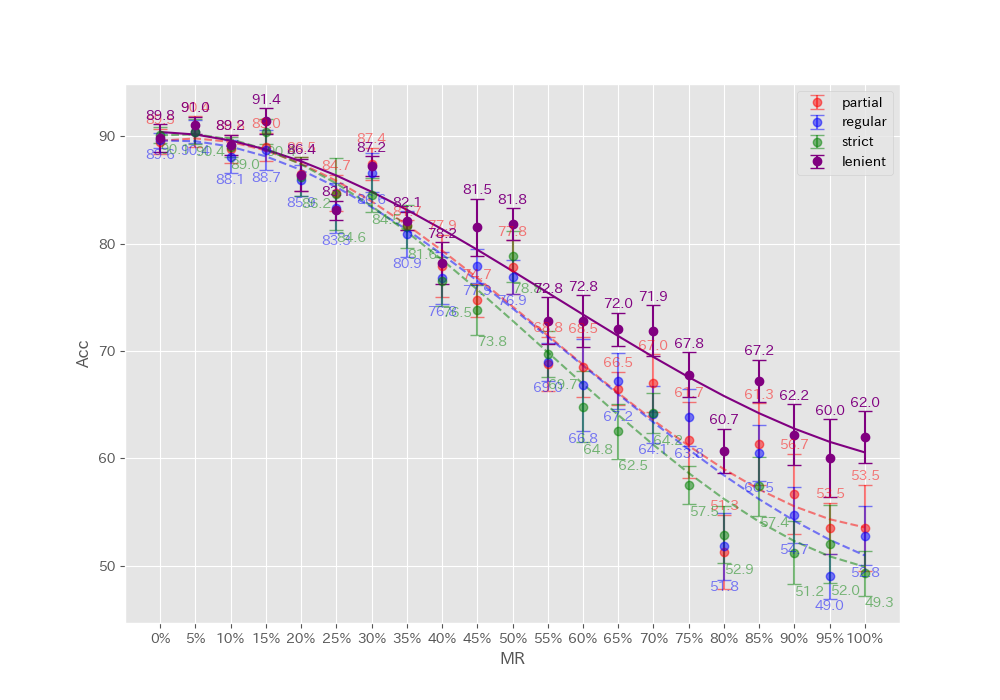}
\end{minipage}
\hspace{-0.09\linewidth}
\begin{minipage}[t]{0.62\linewidth}
        \includegraphics[width=0.95\linewidth]{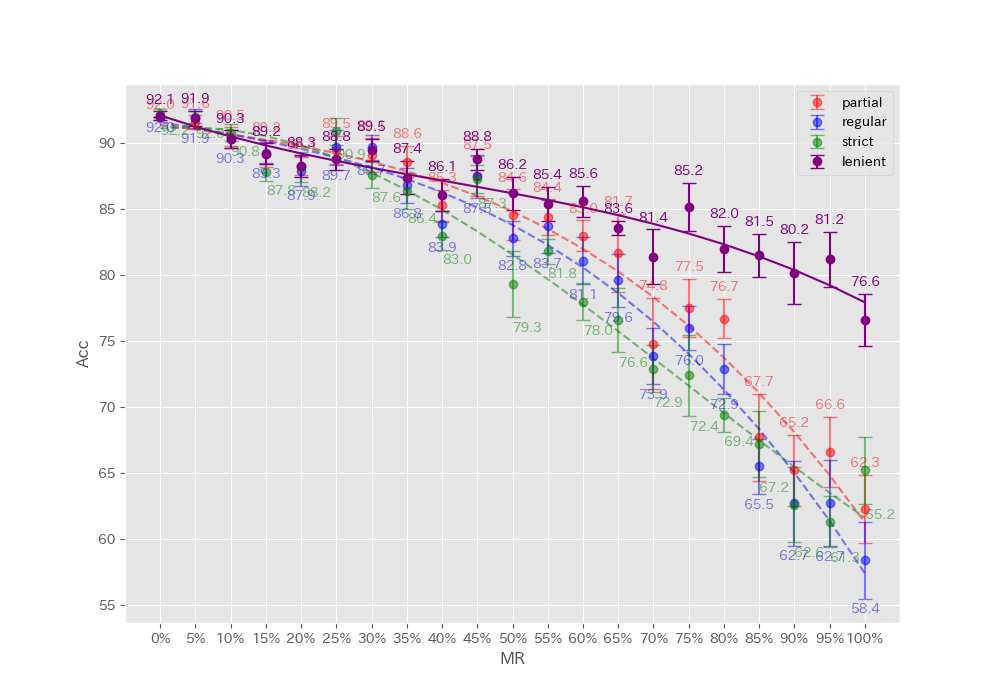}
	\end{minipage}
\captionsetup{width=.95\linewidth}
\caption{Accuracy curves for GPT-4o-mini with four masking methods (strict, regular, partial, and lenient). Left: RQA dataset results. Right: UQA dataset showing higher overall accuracy, particularly with lenient masking.}
\label{fig:acc_curve}
\end{figure}

Figure \ref{fig:acc_curve} (left panel) shows a comparison of lenient/partial/regular/strict maskings for RQA using GPT-4o-mini. The masking rate varies in 5\% increments, with the question-answering process repeated 10 times and average Acc values plotted. As masking rate increases, differences become more apparent. The results indicate that avoiding verb masking effectively maintains high accuracy levels, with this effect particularly pronounced in UQA as shown in the right panel of Figure \ref{fig:acc_curve}.

The right panel of Figure \ref{fig:acc_curve} presents similar verification results for UQA. A notable difference from RQA is that high accuracy is achieved simply by masking content words while excluding verbs. Although background knowledge presence/absence would expectedly lead to such significant differences, this outcome is quite intriguing. For the strict masking curve (green), the medium MR range (45-80\%) shows a clear accuracy drop in the order of partial, regular, and strict masking. However, in the higher MR range (85\% or more), the decline in Acc seems to stabilize, and it's interesting to note that performance becomes equivalent to partial lifting. The similar tendency can be observed in the left panel of Figure~\ref{fig:acc_curve}.

%
%
\vspace{-0.3cm}

%
%
\subsection{Another language model}\label{sec:llama}
\vspace{-1em}

\begin{figure}[h]
\hspace{-0.07\linewidth}
\begin{minipage}[t]{0.62\linewidth}
     \includegraphics[width=0.95\linewidth]{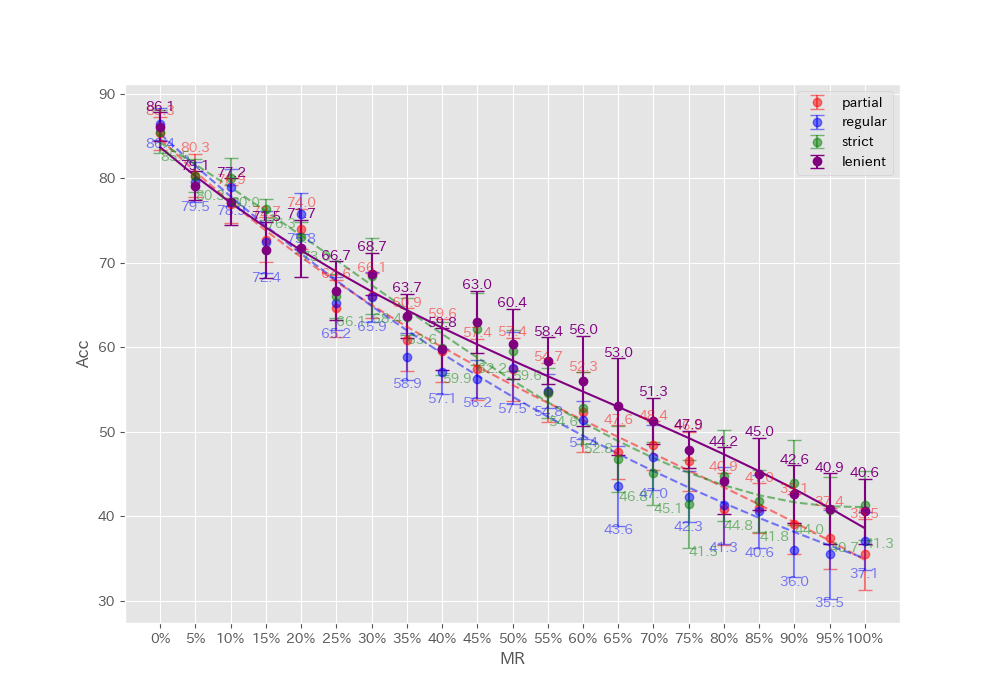}
\end{minipage}
\hspace{-0.09\linewidth}
\begin{minipage}[t]{0.62\linewidth}
    \includegraphics[width=0.95\linewidth]{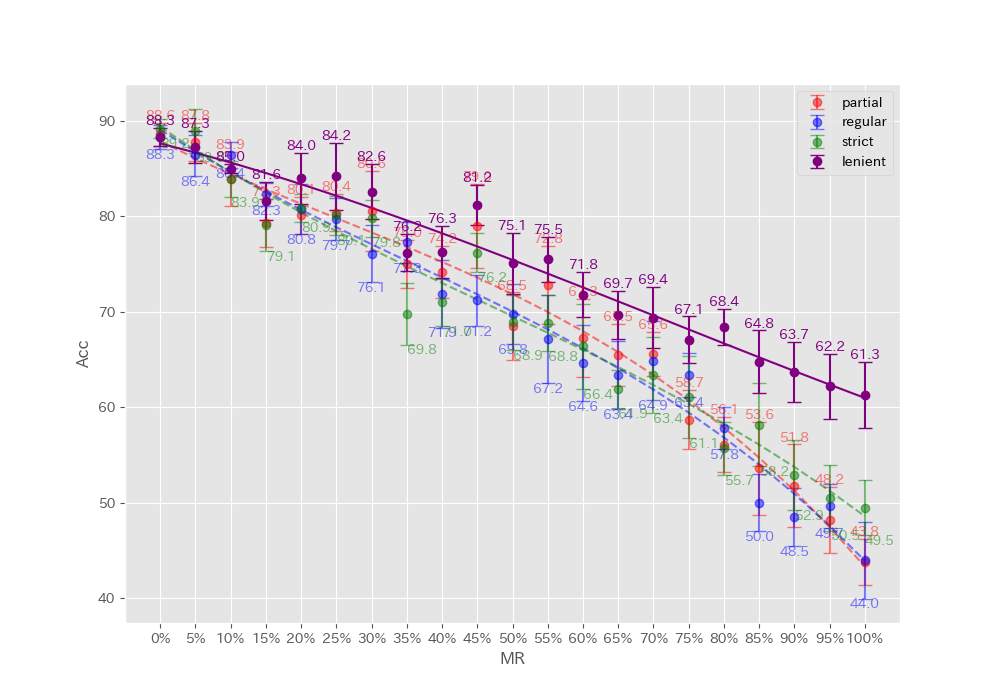}
\end{minipage}
 \captionsetup{width=0.95\linewidth}
 \caption{Accuracy curves for Llama3.2. Left: RQA results showing concave downward trend. Right: UQA results demonstrating similar pattern with higher baseline accuracy.}
     \label{fig:uqa-llama}
 \vspace{-0.5cm}
\end{figure}

We also present Llama3.2 (llama-3.2-3b-instruction) results for comparison as another lightweight LM example. Figure \ref{fig:uqa-llama} (left panel) shows that RQA Acc curves are concave downward, indicating lower mask resilience compared to GPT-4o-mini. The right panel of Figure \ref{fig:uqa-llama} shows similar results for UQA.
\vspace{-0.3cm}

%
\subsection{Arithmetic calculations: AQuA-RAT}\label{sec:math}
\vspace{-1em}

To assess whether computational reasoning can be performed effectively in masked states, neither answer choices nor numbers directly affecting calculation results are masked. Evidence sentence verification (thought process sample sentences) follows three settings: (1) using the same numbers as the question without providing the correct answer, (2) using different numbers from the question and including the correct answer, and (3) providing no thought process (i.e., the same setting as RQA).

\begin{figure}[h]
\hspace{-0.07\linewidth}
\begin{minipage}[t]{0.62\linewidth}
     \includegraphics[width=0.95\linewidth]{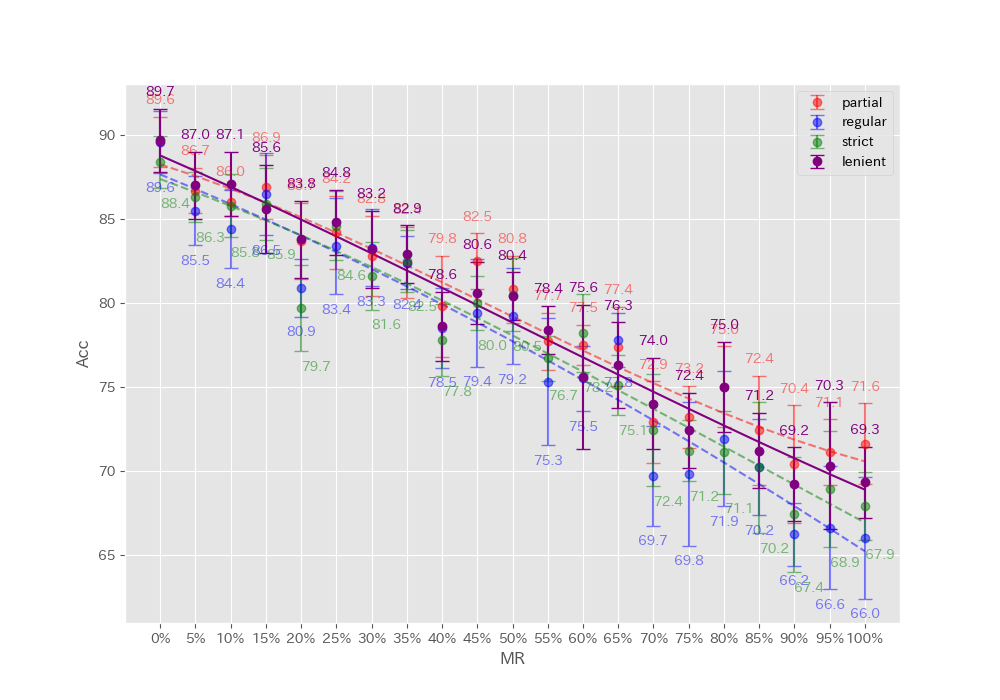}
\end{minipage}
\hspace{-0.09\linewidth}
\begin{minipage}[t]{0.62\linewidth}
    \includegraphics[width=0.95\linewidth]{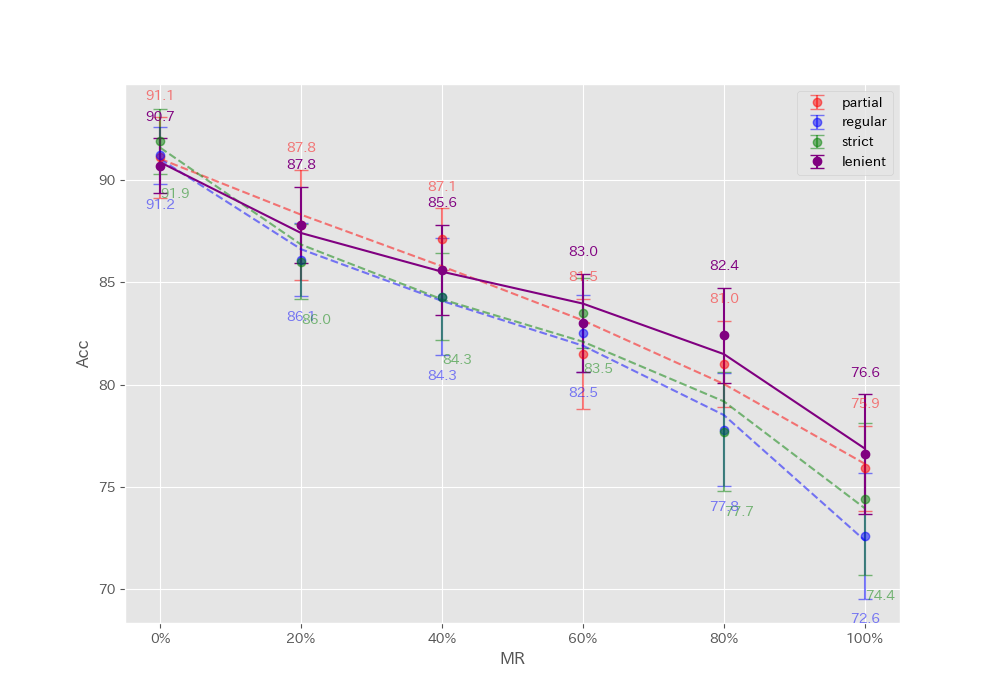}
\end{minipage}
   \captionsetup{width=0.95\linewidth}
    \caption{AQA accuracy curves for Case 1 (math problems with thought process). Left: GPT-4o-mini results. Right: GPT-4o results showing higher overall accuracy.}
     \label{fig:aqua1}
\end{figure}

{\bf Case 1}: Refer to Figure~\ref{fig:aqua1}. Clear trends distinct from UQA and RQA can be observed: (1) Strict masking achieves higher accuracy than regular masking; (2) Lenient masking does not surpass partial lifting accuracy; (3) The response curve behaves linearly. These results likely reflect math problem text characteristics, with fewer words and limited vocabulary variation. This can also be partially observed in GPT-4o results shown in the right side of Figure~\ref{fig:aqua1}.
{\bf Case 2}: 
None of the masking rates achieved accuracy above 40\%. Additionally, issues arose with numerical transformation consistency. Hence, further investigation into this case will not be pursued at this time.
\vspace{-0.3cm}

\begin{figure}[h]
\hspace{-0.07\linewidth}
\begin{minipage}[t]{0.62\linewidth}
\includegraphics[width=0.95\linewidth]{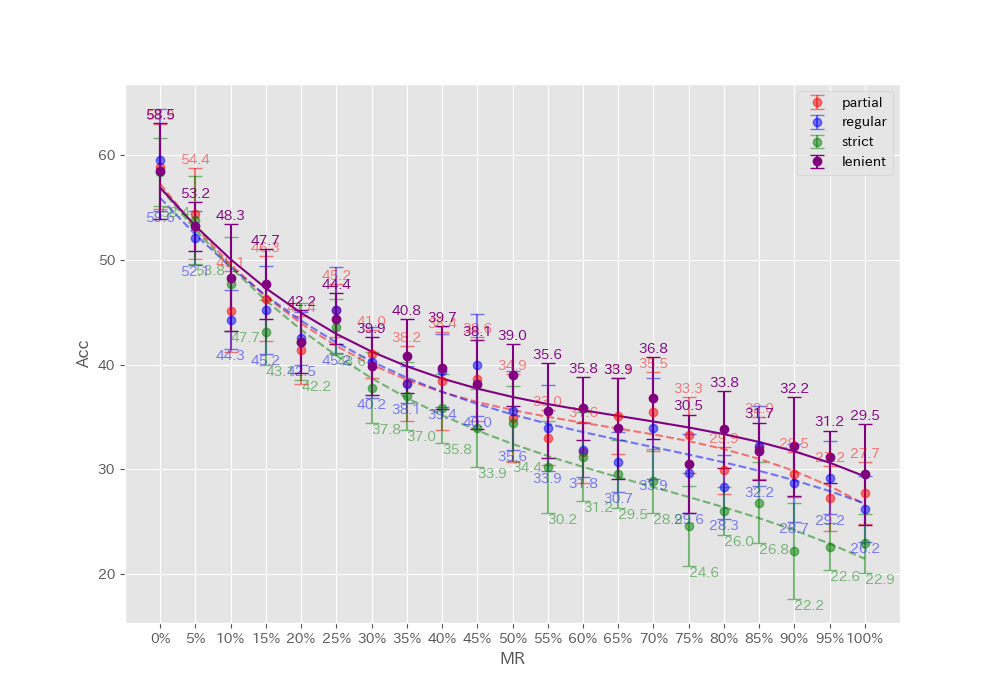}
\end{minipage}
\hspace{-0.09\linewidth}
\begin{minipage}[t]{0.62\linewidth}
\includegraphics[width=0.95\linewidth]{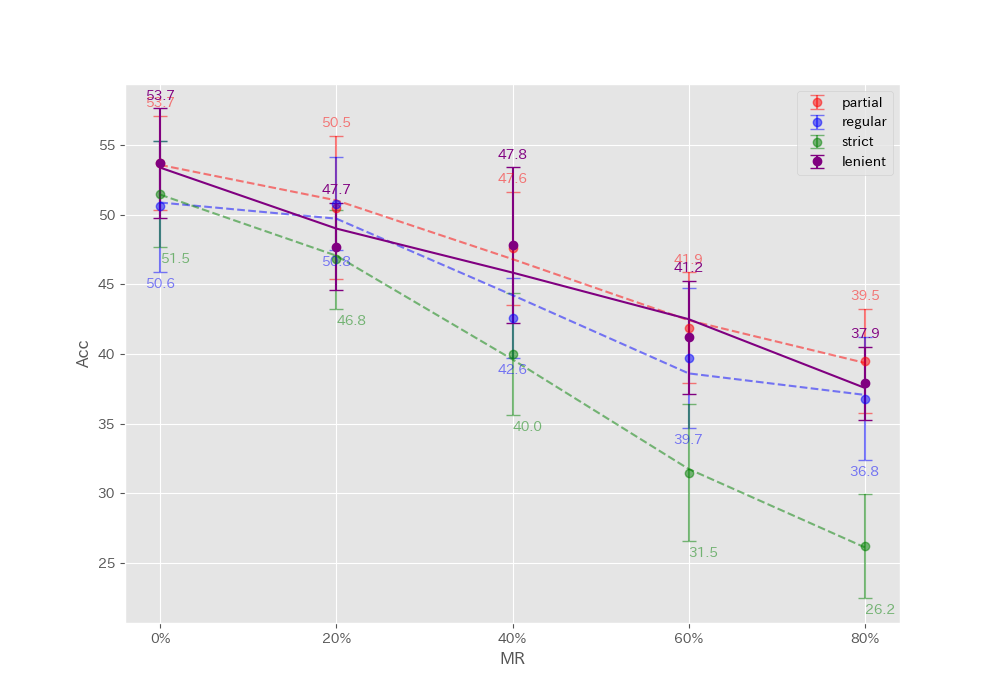}
\end{minipage}
\captionsetup{width=0.95\linewidth}
\caption{AQA accuracy curves for Case 3 (question-only masking without rationale). Left: GPT-4o-mini showing steeper accuracy decline. Right: GPT-4o demonstrating greater resilience to masking, particularly at moderate masking rates.}
\label{fig:aqua3}
\end{figure}

{\bf Case 3}: Figure \ref{fig:aqua3} illustrates cases where only the question is masked without providing rationale. Words containing mathematical symbols ($+, -, /, \ast$) and single-character variables are excluded from masking. In both cases, accuracy achieved is roughly 50\% of that observed in Case 1. Nevertheless, accuracy curves certainly exceed 20\% --- the theoretical accuracy for random responses in five-choice questions --- even with completely masked text (MR=100\%). This unexpected performance suggests some form of systematic reasoning persists even without explicit semantic information.

%
\vspace{-0.3cm}

%
\subsection{Guided response: MskCal}\label{sec:recall}
\begin{figure}[h]
\centering
\begin{minipage}[t]{0.47\linewidth}
\includegraphics[width=1.3\linewidth]{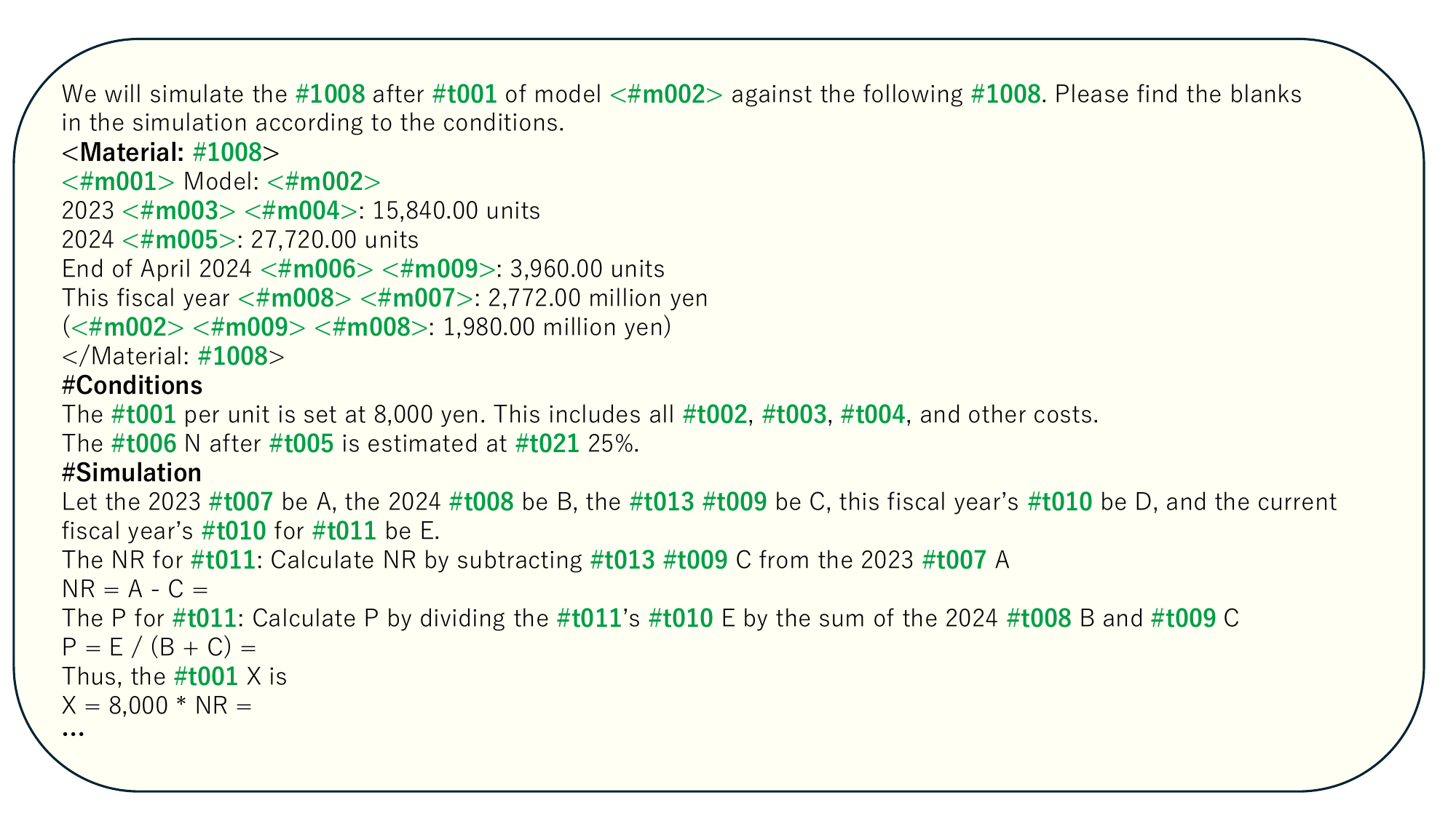}
\end{minipage}
\hfill
\hspace{0.05\linewidth}
\begin{minipage}[t]{0.46\linewidth}
\raggedleft
\includegraphics[width=0.8\linewidth]{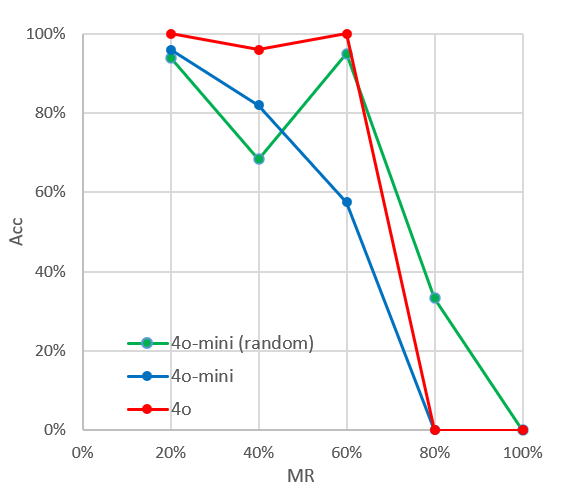}
\end{minipage}
 \captionsetup{width=0.95\linewidth}
\caption{MskCal task examples. Left: An example prompt with masked calculation instructions. Right: Conditional accuracy plots showing performance across different masking rates for GPT-4o and GPT-4o-mini.}
\label{fig:mskcal_acc}
\end{figure}

\vspace{-1em}
In MskCal, we evaluate answer accuracy for written responses rather than multiple-choice questions. We prepare prompts for simple accounting calculations, guiding models to solve problems step by step (Figure~\ref{fig:mskcal_acc}) and assess error in responses to masked guiding sentences. Using a guided format allows us to evaluate logical reasoning ability by substituting it with contextual understanding, assuming logic accuracy will be reflected in numerical answers. Specifically, we measure relative error across ten trials in five variables: sales price ($P$), post-recall sales units ($N$), sales decline ($Y$), corrected sales for the target product ($E'$), and corrected annual sales ($D'$).
%
%

%
%
\begin{table}[h]
\centering
\caption{Performance comparison of masked calculation tasks across different masking rates. Left columns show GPT-4o-mini results for both random and restricted masking conditions, while right columns show GPT-4o results with restricted masking.}
\label{tab:mskcal_acc}
{\renewcommand{\arraystretch}{1.2}
\begin{adjustbox}{max width=\linewidth}
\begin{tabular}{|l|c|c|c|c|c|c|c|c|c|c|c|c|} \hline
& \multicolumn{8}{c|}{GPT-4o-mini} & \multicolumn{4}{c|}{GPT-4o} \\ \hline
& \multicolumn{4}{c|}{Random Masking} & \multicolumn{4}{c|}{Restricted Masking} & \multicolumn{4}{c|}{Restricted Masking} \\ \hline
MR & NAR & ave.$P_\delta$ & ave.$P_\sigma$ & ave.$P_{\sigma/2}$ & NAR & ave.$P_\delta$ & ave.$P_\sigma$ & ave.$P_{\sigma/2}$ & NAR & ave.$P_\delta$ & ave.$P_\sigma$ & ave.$P_{\sigma/2}$ \\ \hline
20\% & 0 & 98.9\% & 94.0\% & 90.0\% & 0 & -2398\% & 96.0\% & 96.0\% & 0 & 99.99\% & 100\% & 100\% \\ \hline
40\% & 0.62 & -442.4\% & 68.3\% & 68.3\% & 0 & -4889\% & 82.0\% & 78.0\% & 0 & 100\% & 96.0\% & 94.0\% \\ \hline
60\% & 0.76 & 79.1\% & 95.0\% & 95.0\% & 0.2 & -3251\% & 57.5\% & 52.5\% & 0.08 & 99.97\% & 100\% & 100\% \\ \hline
80\% & 0.74 & -2141\% & 33.3\% & 33.3\% & 1.0 & 0.0\% & 0.0\% & 0.0\% & 1.0 & 0.0\% & 0.0\% & 0.0\% \\ \hline
100\% & 1.0 & 0.0\% & 0.0\% & 0.0\% & 1.0 & 0.0\% & 0.0\% & 0.0\% & 1.0 & 0.0\% & 0.0\% & 0.0\% \\ \hline
\end{tabular}
\end{adjustbox}
}
\end{table}

Table \ref{tab:mskcal_acc} summarizes missing response rates (NAR) and average success indicators ($P_\delta$, $P_{\nu\sigma}$) for each masking rate. 
These indicators are defined as follows:
$\mbox{ave.}\delta X$ expresses the mean relative error $\delta X$ over 10 trials; 
$P_\delta:=1-\delta\tilde{X}$, where $\delta\tilde{X}$ is trimmed mean of $\delta X$; 
$P_{\nu\sigma}$ is the success rate given by the average over 
$\{\delta X| \delta X\leq\nu31.73\%(=\nu\sigma)\}$. 
Note that $\sigma$ is not the standard deviation of the 10 trials but a constant. The averages represent 5-variable means and conditional success rates when responses were obtained. 

At 20\% masking rate, GPT-4o-mini shows nearly perfect accuracy. However, GPT-4o-mini can produce order-of-magnitude errors (missing zeros), resulting in relative errors of $\mathcal{O}(10^3)$\%. At 40\% masking rate, missing responses increase to 62\%, and even when responses are obtained, accuracy drops below 70\%.

The right side of Figure \ref{fig:mskcal_acc} shows the MR response of $P_\sigma$. The green line represents $P_\sigma$ for GPT-4o-mini with random masking (left columns of Table \ref{tab:mskcal_acc}), while other lines show results for GPT-4o-mini and GPT-4o with restricted masking (middle and right colums of Table \ref{tab:mskcal_acc}, respectively) where computational derivation sections remain unmasked. As shown in the random masking data (left columns of Table \ref{tab:mskcal_acc}), GPT-4o-mini exhibits digit placement errors ($P_\delta$ of O($10^3$)\%) at 20\%-40\% masking rates, with accuracy dropping to 50\% above 60\% masking rate. GPT-4o shows no digit placement errors and maintains favorable results up to 60\% masking rate. However, at masking rates of 80\% or higher, no responses could be generated in any case, as task instruction sections in the prompt are masked, causing both models to fail in decoding.

%
%
\subsection{Indicator Analysis (GPT-4o-mini)}\label{sec:index}
\vspace{-1em}

\begin{figure*}[t]
    \centering
    \captionsetup{width=.95\linewidth}
    \includegraphics[width=.96\linewidth]{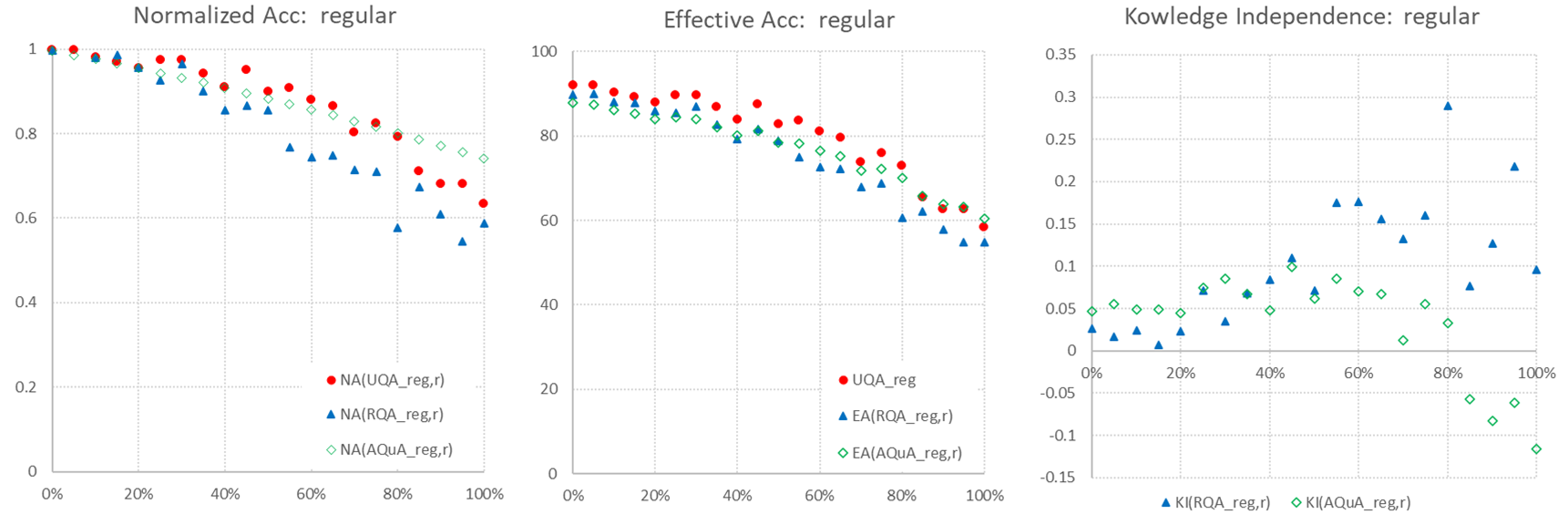}
    \vspace{-0.1cm}
    \caption{Comparisons of effective Acc and knowledge independence (KI). }
    \label{fig:index}
    \vspace{-0.3cm}
\end{figure*}

By considering both source characteristics and accuracy metrics, we analyze LLM performance in the decoding task. Figure \ref{fig:index} compares NA, EA and KI metrics for regular masking across different sources. For AQA, Case 1 results are used.

First, examining NA (normalized by mask-free accuracy), AQA shows results as high as UQA, while RQA is relatively lower. From accuracy degradation perspective relative to mask-free state, this suggests AQA maintains higher performance even under masked conditions, indicating logical reasoning remains possible despite masking. However, as shown in Figure \ref{fig:mskcal_acc}, MskCal accuracy, associated with computational capability, is highly sensitive to masking rate changes. Thus, AQA's high NA value likely indicates strong decoding ability, discounted for computational limitations.

Second, our KI observations show AQA exhibits lower values than RQA except at masking rates $\lesssim40\%$. While higher masking rates should theoretically reduce knowledge dependency and increase KI --- a trend clearly observed in RQA's KI graph (Figure \ref{fig:index}) --- AQA shows a different pattern. In arithmetic problems like AQA, where background knowledge should minimally impact answers, KI values are expected to be naturally higher than usual. Although AQA shows higher KI than RQA at masking rates $\lesssim40\%$, it declines after 60\%. This performance degradation beyond 60\% aligns with trends observed in MskCal, suggesting computational reasoning under masked conditions may be more challenging than contextual inference. For further detailed analysis, see Appendix \ref{sec:D}.

\blue{Although this study does not explicitly separate the roles of syntax, semantics, and memorization in LLM reasoning, we note that our masking variants indirectly reflect these aspects. For instance, strict masking eliminates semantic cues and thus increases reliance on memorized patterns or abstract inference; lenient masking preserves syntactic scaffolding by retaining verbs; and comparisons with UQA highlight the influence of background knowledge. A more targeted decomposition of these factors is an important direction for future work.}

\blue{
While our current results primarily focus on accuracy curves, deeper analysis of error types and reasoning failure modes will be necessary in future work to better understand LLM behavior under masking.}

%
%
\section{Conclusion and Discussion}\label{sec:concl}
In this study, we proposed a data preparation method and evaluation metrics to measure LLMs' ability to perform question-answering tasks on masked text. We built a masked dataset and conducted evaluations using it. Our defined evaluation metrics (NA, EA, KI) account for differences between cases where background knowledge is available during decoding, aiming to neutralize or remove knowledge dependency influence. The question-answering tasks were validated in two formats: multiple-choice (MskQA) and descriptive computational reasoning (MskCal). For MskQA, we used masked datasets based on RQA and AQA, while for MskCal, we employed accounting calculation problems we created.

Our results show that model answering ability is highly dependent on masking rate. For MskQA, GPT-4o-mini demonstrates accuracy of Acc $\gtrsim60\%$ (50\% for RQA, 70\% for UQA and AQA) even with masking rates over 80\%. To maintain accuracy around 80\% or higher, masking rates need to be kept below approximately 60\% (30\% for RQA, 70\% for UQA, 50\% for AQA), and below 40\% for MskCal.

For GPT-4o, the maximum masking rate maintaining 80\% accuracy is MR$\lesssim70\%$ for MskQA (AQA) and MR$\lesssim60\%$ for MskCal, demonstrating significantly higher masking rate tolerance compared to GPT-4o-mini. In both models, masking rates of MR$\lesssim40\%$ generally achieve around 80\% average accuracy across question formats. Despite substantial tokenization changes caused by masked text, powerful LLMs like GPT-4o and 4o-mini demonstrate effective processing of various masked text levels. However, both models struggle with heavily masked computational reasoning tasks (masking rates MR$\gtrsim60\%$).

Notably, GPT-4o shows superior performance in computational reasoning tasks with MskCal, significantly outperforming 4o-mini. Additionally, in masked question-answering tasks, GPT-4o maintains very high accuracy with masked contexts, achieving results closely matching its original accuracy. 

%
\blue{
Although this study focused on general-purpose QA and arithmetic tasks, the proposed framework is not limited to them. It could be extended to domain-specific settings (e.g., legal, biomedical, scientific QA), which would allow further analysis of how LLMs process masked information in specialized knowledge environments. The current dataset selection was a deliberate choice to isolate inference behavior and minimize the effects of memorization, particularly by leveraging post-training sources such as RealtimeQA.}

While various word-masking methods exist (e.g., category abstraction, synonym replacement), this study focuses solely on semantic withholding of content words and morpheme limitation. Future work could explore more complex scenarios including function word masking.

Our experiments are limited to two datasets (RealtimeQA and AQuA-RAT) and a custom MskCal task for sales impact calculations. While MskCal was specifically designed to isolate contextual understanding from computational ability, extending this approach to diverse datasets remains challenging, despite the general applicability of both MskQA and MskCal frameworks. Another issue concerns masked information generation. For instance, in computational reasoning problems, certain LLMs tend to make digit-related mistakes, which may be exacerbated by masking --- resulting in further errors such as misinterpreting units. These tendencies might be controlled by adjusting meta-information construction.

In this paper, we examined various LLMs' abilities to process masked text under different settings. However, understanding why some LLMs, particularly GPT-4, excel at these tasks remains challenging due to limited access to closed-source models. Testing existing hypotheses about training methods and model scaling could offer valuable insights into LLMs' inner workings.

\blue{
We did not include a human performance baseline, as the primary focus of this study is the behavior of LLMs under masked semantic constraints. Given that the masked inputs consist of abstract code labels and sparse context, as shown in Appendices \ref{sec:B}  and \ref{sec:C} as well as in GitHub examples, we believe these tasks are beyond typical human interpretability. Human benchmarking, while potentially interesting, falls outside the technical scope of this work.}

\appendix
\setcounter{equation}{0}
\section{Generation of UQA dataset}
\label{sec:A}
\indent
We generated the UQA dataset by creating question-answer pairs based on widely available foundational documents such as the Declaration of Independence (1776), the Gettysburg Address (1863), the UN Charter (1945), the UNESCO Constitution (1945), the Universal Declaration of Human Rights (1948), and the Convention on the Rights of the Child (1989). These documents were selected for their clear language, universal availability, and content that is unlikely to have been specifically optimized for in LLM training.

\subsection{Question Generation}
\vspace{-1em}
We used the following prompt to generate multiple-choice questions from these documents:
\par
\setlength{\parskip}{0pt}
\vspace{3mm}
\noindent
{\tt
Based on the following text, create three simple multiple-choice questions that can be answered by a middle school student. Ensure that the questions do not 
require any background knowledge and can be answered using only the information provided in the text. Present the questions in the following format: \par
[Question text] \par
A) [Option A]\par
B) [Option B]\par
C) [Option C]\par
Answer: [Correct option]\par
Text: [Insert the provided text here]
}
\vspace{0.5cm}

\noindent
A sample of the original QA generated by Llama3-70b-instruct:\par
\begin{spverbatim}
Question: What is the main purpose of the Declaration of Independence, according to the text?
Choices: (A)To establish a new government (B)To elect a new president (C)To explain why a separation is necessary (D)To declare war on another country
Text: When in the Course of human events, it becomes necessary for one people to dissolve the political bands which have connected them with another, and to assume among the powers of the earth, the separate and equal station ...(Omitted)
----
(C) To explain why a separation is necessary
\end{spverbatim}


\subsection{Example of masked prompt}\label{sec:A_uqa}

Masked prompt for the above QA (MR 100\%, meta-information partially shown due to space limitations):\par
\begin{spverbatim}
The following is a text and metadata related to the code terms within the text. Answer the question concisely according to the instructions.
## Instructions
- Choose the answer from the options and respond with the corresponding number.
- Respond in JSON format as {'basis': str, 'answer': int}
- Use only the text as a reference for the basis
## Text
When in the <r009> of <r021> <r018>, it <r005> <r025> for one <r029> to <r012> the <r030> <r004> which have <r007> them with another, and to <r003> ...(Omitted)
## Question
What is the <r023> <r033> of <r040>, <r002> to the <r039>?
## Options
['1. To <r017> a <r026> <r020>', '2. To <r014> a <r026> <r032>', '3. To <r019> why a <r037> is <r025>', '4. To <r011> <r042> on another <r008>']
## Metadata
part_of_speech | category | meaning | code
PROPN | Individual Name | Higher power | r001
VERB |  | Agreement | r002
VERB |  | Take on, responsibility | r003
NOUN | Organization Name | Group of people | r004
(Omitted)
\end{spverbatim}

\setcounter{equation}{0}
\section{Masked prompting of MskQA}
\label{sec:B}
\indent
Appendix B presents examples of original QA and thier masked prompts for RQA and AQA.
The masking format for RQA is completely the same as Appendix A. 
Below is the original QA from Figure 1:

\begin{spverbatim}
Question: Which pop duo are embroiled in a legal battle after one obtained 
a restraining order against the other?
Choices: (A)Hall \& Oates  (B)Simon \& Garfunkel  (C)The Righteous Brothers  
(D)The White Stripes
Evidence: Daryl Hall is suing his musical partner John Oates and has obtained a restraining order against him as part of the sealed lawsuit. The case between the US duo, who are the most commercially successful duo in the history of pop music, reportedly involves ``contract/debt''.
Answer: Bsed on the evidence, among A through D, the answer is
----
(A)Hall \& Oates
\end{spverbatim}

\vskip\baselineskip
\subsection{AQuA-RAT (AQA)}\label{sec:B_aqua}

We show an example of AQA's original QA and illustrate masked prompts (MR 100\%).
The thought process is given in Rationales, and the CoT prompt is structured as \cite{Wei2022b}. Here, we provide an example of the Case 1 format in which the final part of the Rationale does not contain the answer selection, but only the process, and gives the same numerical value as the question. The riginal QA is:
\medskip\noindent
\begin{spverbatim}
Rationale: Total weight of the 4 dogs = (25+31+43+41) = 140
Avg = 140/4 = 35
Total weight of 5 dogs = 140+y or 4(35)+y
Average of 5 dogs as per Question = 35
Equation:
4(35) + y = 5 (35), or y = 35.
Question: The weight of four dogs is determined to be 25 pounds, 31 pounds, 43 pounds and 41 pounds respectively. The weight of a fifth dog is determined to be y pounds. If the average (arithmetic mean) weight of the first four dogs is the same as that of all five dogs what is the value of y?
Options: (A)31 (B)33 (C)35 (D)37 (E)39
----
Choose (C) 
\end{spverbatim}

\medskip\noindent
\subsection{Masked prompt: the Case 1 AQA (MR 100\%)}
\begin{spverbatim}
The following is a text and metadata related to the code terms within the text. Answer the question concisely according to the instructions.
## Instructions
- Choose the answer from the options and respond with the corresponding number.
- Respond in JSON format as {'basis': str, 'answer': int}
- Use only the text as a reference for the basis
## Text (Rationale)
<r016> <r018> of the 4 <r009> <r001> (25+31+43+41) <r001> 140
<r003> <r001> 140/4 <r001> 35
<r016> <r018> of 5 <r009> <r001> 140+<r019> or 4(35)+<r019>
Average of 5 <r009> as per <r013> <r001> 35
<r010>:
4(35) + <r019> <r001> 5 (35), or <r019> <r001> 35.
## Question
The <r018> of four <r009> is <r008> to be 25 <r012>, 31 <r012>, 43 <r012> and 41 <r012> respectivel<r019>. The <r018> of a fifth <r009> is <r008> to be <r019> <r012>. If ...(Omitted)... what is the <r017> of <r019>?
## Options
['1. 31', '2. 33', '3. 35', '4. 37', '5. 39']
## Metadata
part_of_speech | category | meaning | code
VERB, NOUN |  | comparison, equation symbol | r001
PROPN | GPE | average value | r003
ADJ |  | mathematical operation | r004
(Omitted)
\end{spverbatim}

\setcounter{equation}{0}
\section{MskCal Prompt}
\label{sec:C}
\indent
Appendix C summarizes the original questions and a sample of masked prompt with masking codes utilized in MskCal verification. Correct answers and mask-free results are given in Table~\ref{tb:error_comparison}. 

\subsection{Original questions}
\begin{spverbatim}
We will simulate the sales plan after the recall of the ZX-1000 model based on the following sales plan. Please fill in the blanks in the simulation according to the conditions.
#Document: Sales Plan
Scooter Model: ZX-1000
2023 Production Volume: 15,840.00 units
2024 Production Plan: 27,720.00 units
Domestic Inventory as of the end of April 2024: 3,960.00 units
Projected Revenue for This Fiscal Year: 2,772.00 million yen
(ZX-1000 Domestic Projected Revenue: 1,980.00 million yen)
#Conditions
The recall cost per unit is set at 8,000 yen, which includes all costs such as parts, repairs, transportation, and other expenses.
The post-recall sales volume N is estimated with a reduction rate of 25%.
#Simulation
Let A be the production volume in 2023, B the production plan volume for 2024, C the inventory volume as of April 2024, D the planned revenue for this fiscal year, and E the planned revenue for this fiscal year for the model subject to recall. The number of units sold subject to recall, NR: calculated by subtracting the number of units remaining unsold as of April 2024 from the 2023 production volume A, i.e., NR = A - C =
The sales price of the model subject to recall, P: calculated by dividing the planned sales revenue E of the model by the total of the production plan volume B and the inventory volume C for 2024, i.e., P = E / (B + C) =
Therefore, the total recall cost X is, X = 8,000 * NR =
Since the planned sales volume is B + C, considering the reduction rate, 
the post-recall sales volume N is, N = (B + C) * (1 - 0.25) =
The decrease in revenue Y is, Y = P * (B + C) * 0.25 =
The loss amount L is, L = X + Y =
The revised planned sales revenue for the model subject to recall, E', is, 
E' = E - L =
The revised planned revenue for this fiscal year, D', is, 
D' = D - L =
\end{spverbatim}

\subsection{Masked questions}
The masked prompt of the guided question is as follows (MR 20\%, some parts are omitted for simplicity):\par
\begin{spverbatim}
We will simulate the #m010 after #t001 of model <#m002> against the following #m010. Please find the blanks in the simulation according to the conditions.
#Document: #m010
<#m001> Model: <#m002>
2023 <#m003> <#m004>: 15,840.00 units
2024 <#m005>: 27,720.00 units
End of April 2024 <#m006> <#m009>: 3,960.00 units
This fiscal year <#m008> <#m007>: 2,772.00 million yen
(<#m002> <#m009> <#m008>: 1,980.00 million yen)
#Conditions
The #t001 per unit is set at 8,000 yen. This includes all #t002, #t003, #t004, and other costs. The #t006 N after #t005 is estimated at #t021 25%.
#Simulation
Let the 2023 #t007 be A, the 2024 #t008 be B, the #t013 #t009 be C, this fiscal year’s #t010 be D, and the current fiscal year’s #t010 for #t011 be E. The NR for #t011: Calculate NR by subtracting #t013 #t009 C from the 2023 #t007 A 
NR = A - C =
The P for #t011: Calculate P by dividing the #t011’s #t010 E by the sum of the 2024 #t008 B and #t009 C for 2024, i.e., P = E / (B + C) =
Therefore, the #t001 X is X = 8,000 * NR =
(Omitted)
<Meta Information: Document>
Number Part of Speech Category Meaning
m001 General Product Transportation, Vehicle
m002 Proper Noun Product Product Name
m003 General Planning Manufacturing, Creation
(Omited)
\end{spverbatim}

%
%
\begin{table}[h]
\centering
\caption{Comparison of error metrics at 0\% masking rate between GPT-4o-mini and GPT-4o, showing the superior numerical precision of GPT-4o across all variables.}
\label{tb:error_comparison}
{\renewcommand{\arraystretch}{1.2}
\begin{adjustbox}{max width=\linewidth}
\begin{tabular}{|l|l|c|c|c|c|c|c|c|c|c|} \hline
\multirow{2}{*}{Variable} & \multirow{2}{*}{True value} & \multicolumn{4}{c|}{GPT-4o-mini} & \multicolumn{4}{c|}{GPT-4o} \\ \cline{3-10}
 & & ave.$\delta X$ & $P_\delta$ & $P_\sigma$ & $P_{\sigma/2}$ & ave.$\delta X$ & $P_\delta$ & $P_\sigma$ & $P_{\sigma/2}$ \\ \hline
P & 62,500 & 20.0\% & 87.5\% & 80.0\% & 80.0\% & 0.00\% & 100.0\% & 100.0\% & 100.0\% \\ \hline
N & 23,760 & 0.00\% & 100\% & 100\% & 100\% & 0.00\% & 100.0\% & 100.0\% & 100.0\% \\ \hline
Y & $4.95\times10^8$ & 10.2\% & 99.7\% & 90.0\% & 90.0\% & 5.03\% & 99.97\% & 90.0\% & 90.0\% \\ \hline
E' & $1.38996\times10^9$ & 3.68\% & 99.9\% & 90.0\% & 90.0\% & 1.79\% & 99.99\% & 100\% & 90.0\% \\ \hline
D' & $2.18196\times10^9$ & 2.35\% & 99.9\% & 100\% & 90.0\% & 1.14\% & 99.99\% & 100\% & 100\% \\ \hline
Average & - & 7.26\% & 97.4\% & 92.0\% & 90.0\% & 1.59\% & 99.99\% & 98.0\% & 96.0\% \\ \hline
\end{tabular}
\end{adjustbox}
}
\end{table}

\setcounter{equation}{0}
\section{Details of indicator analysis}
\label{sec:D}
\indent
This appendix presents detailed analysis of Section~\ref{sec:index} for GPT-4o-mini. Table~\ref{tab:Acc_mean} compares Acc across four masking levels and the metrics NA, EA, and KI for regular masking across different sources, with Case 1 results used for AQA. 
Instead of visually comparing Acc response curves for each dataset, we introduce indices for quantitative comparison by averaging with respect to the weight of $r$. We define two types of $r$-weighted averages for each indicator $X=NA, EA, KI$:
\[
X_1(D)=\frac{1}{k_1}\sum_r r\ast X(D,r)\,,\quad  
X_2(D)=\left( \prod_{r} X(D,r) \right)^{\frac{1}{k_2}}\,,  
\]
where $k_1=\sum_r r$ and $k_2=\sum_r 1$.

Acc increases across all sources as masking level loosens from strict to lenient, though AQA shows some fluctuations. AQA exhibits lower Acc compared to RQA, likely due to its computational elements requiring complex calculations under masked conditions. 
In contrast, AQA's KI plateaus at MR=40\% and declines after 60\% (Figure~\ref{fig:index}), consistent with MskCal's performance degradation beyond 60\% masking rate (Figure~\ref{fig:mskcal_acc}). This pattern explains AQA's smaller $r$-weighted average KI compared to RQA.

Interestingly, the four-state average of $X_i(Acc)$ (denoted as $\langle X_i(Acc) \rangle$) closely matches the $X_i$ average of EA (Table~\ref{tab:Acc_mean}), suggesting EA can be considered effective Acc. However, AQA remains exceptional, showing a notable discrepancy between Acc and EA values. The fact that AQA's EA exceeds RQA's remains an open question requiring further investigation.

%
%
\begin{table}[ht]
\centering
\caption{Comprehensive comparison of accuracy metrics across masking techniques for UQA, RQA, and AQA datasets.}
\label{tab:Acc_mean}
{\renewcommand{\arraystretch}{1.2}
\begin{adjustbox}{max width=\linewidth}
\begin{tabular}{|c|c|c|c|c|c|c|c|c|c|c|}
\hline
\multirow{2}{*}{\begin{tabular}{c}Dataset \\ (mask-free Acc)\end{tabular}} & \multirow{2}{*}{Metric} & \multicolumn{4}{c|}{Masking Type (Acc)} & \multicolumn{3}{c|}{Regular Masking Metrics} & \multicolumn{2}{c|}{Four-state Average} \\ \cline{3-11}
 &  & strict & reg. & part. & lenient & NA & EA & KI & $< X_i(Acc) >$ & $X_i(EA)$ \\ \hline
\multirow{2}{*}{\begin{tabular}{c}\textbf{UQA} \\ (92.08)\end{tabular}} & $X_1$ & 73.83 & 74.50 & 76.53 & 83.44 & 0.8091 & 74.50 & 0 & 77.07 & 74.50 \\ \cline{2-11}
 & $X_2$ & 79.01 & 79.67 & 81.16 & 85.68 & 0.8653 & 79.67 & 0 & 81.38 & 79.67 \\ \hline
\multirow{2}{*}{\begin{tabular}{c}\textbf{RQA} \\ (89.84)\end{tabular}} & $X_1$ & 63.27 & 64.67 & 65.48 & 70.60 & 0.7198 & 68.51 & 0.1344 & 66.01 & 68.51 \\ \cline{2-11}
 & $X_2$ & 70.29 & 71.29 & 71.97 & 75.99 & 0.7935 & 74.45 & 0.0722 & 72.39 & 74.45 \\ \hline
\multirow{2}{*}{\begin{tabular}{c}\textbf{AQA} \\ (88.02)\end{tabular}} & $X_1$ & 47.93 & 47.82 & 50.33 & 49.90 & 0.8307 & 72.12 & 0.0093 & 49.00 & 72.12 \\ \cline{2-11}
 & $X_2$ & 51.15 & 50.80 & 52.73 & 52.45 & 0.8741 & 76.55 & 0.0579 & 51.78 & 76.55 \\ \hline
\end{tabular}
\end{adjustbox}
}
\end{table}



\begin{thebibliography}{99}
\bibitem{Islam2024}
Islam R, Moushi OM. 
GPT-4o: The Cutting-Edge Advancement in Multimodal LLM. 
TechRxiv. DOI:10.36227/techrxiv.171986596.65533294. 2024.

\bibitem{Gemini2023}
Gemini Team and et. al. Gemini: A Family of Highly Capable Multimodal Models. 
https://arxiv.org/pdf/2312.11805

\bibitem{Touvron2023}
Touvron H, Lavril T, Izacard G, et. al. 
LLaMA: Open and Efficient Foundation Language Models. 
https://arxiv.org/pdf/2302.13971



\bibitem{Wei2022a}
Wei J, Tay Y, Bommasani R, et. al. Emergent Abilities of Large Language Models. 
Transactions on Machine Learning Research (TMLR Survey Certification). 2022.

\bibitem{Wei2022b}
Wei J, Wang X, Schuurmans D, et. al. Chain-of-thought prompting elicits reasoning in large language models.  
Advances in Neural Information Processing Systems. 2022; 35:24824-24837.

\bibitem{Matsuo2023}
Cao Q, Kojima T, Matsuo Y, Iwasawa Y. 
Unnatural Error Correction: GPT-4 Can Almost Perfectly Handle Unnatural Scrambled Text. 
Proceedings of the 2023 Conference on Empirical Methods in Natural Language Processing (EMNLP). 2023; 8898–8913.

\bibitem{Sinha2021a}
Sinha K, Jia R, Hupkes D, Pineau J, Williams A, Kiela D. 
Masked language modeling and the distributional hypothesis: Order word matters pre-training for little. 
Proceedings of the 2021 Conference on Empirical Methods in Natural Language Processing. 2021; 2888-2913. 

\bibitem{Sinha2021b}
Sinha K, Parthasarathi P, Pineau J, Williams A. UnNatural Language Inference. 
Proceedings of the 59th Annual Meeting of the Association for Computational Linguistics and the 11th International Joint Conference on Natural Language Processing. 
2021;1:7329–7346.

\bibitem{Abdou2022}
Abdou M, Ravishankar V, Kulmizev A, S{\o}gaard A. 
Word order does matter and shuffled language models know it. 
Proceedings of the 60th Annual Meeting of the Association for Computational Linguistics.  2022;1:6907-6919.

\bibitem{Zhao2024} 
Zhao Q, Li J, Li L, Zhou Z, Liu J. 
Word Order's Impacts: Insights from Reordering and Generation Analysis. 
https://arxiv.org/pdf/2403.11473

\bibitem{Salazar2020}
Salazar J, Liang D, Nguyen TQ, Kirchhoff K. 
Masked Language Model Scoring. 
Proceedings of the 58th Annual Meeting of the Association for Computational Linguistics.  2020;2699-2712. 

\bibitem{Song2019}
Song K, Tan X, Qin T, Lu J, Liu TY. 
MASS: Masked Sequence to Sequence Pre-training for Language Generation. 
Proceedings of the 36th International Conference on Machine Learning. 
2019;PMLR 97:5926-5936. 

\bibitem{Sun2019}
Sun K, Yu D, Chen J, Yu D, Choi Y, Cardie C. 
DREAM: A challenge data set and models for dialogue-based reading comprehension. 
Transactions of the Association for Computational Linguistics. 
2019;7:217-231.

\bibitem{OpenAI2023}
OpenAI, et. al. GPT-4 Technical Report. 
https://arxiv.org/pdf/2303.08774

\bibitem{Vaswani2017}
Vaswani A, et. al. Attention is All You Need. https://arxiv.org/pdf/1706.03762

\bibitem{Brown2020}
Brown TB, et al. Language Models are Few-Shot Learners. 
Advances in Neural Information Processing Systems (NeurIPS). 2020;33:1877–1901.

\bibitem{Kasai2022}
Kasai J, Sakaguchi K, Takahashi Y, et. al. RealTime QA: What's the Answer Right Now? 
Advances in Neural Information Processing Systems 36 (NeurIPS 2023) Datasets and Benchmarks Track.

\bibitem{Ling2017}
Ling W, Yogatama D, Dyer C, Blunsom P. 
Program Induction by Rationale Generation : Learning to Solve and Explain Algebraic Word Problems. https://arxiv.org/pdf/1705.04146

\end{thebibliography}
\end{document}